\documentclass[11pt]{article} 
\usepackage{rldmsubmit,palatino}
\usepackage{graphicx}

\title{Modelling bounded rational decision-making through Wasserstein constraints}

\author{
Benjamin Patrick Evans\\
JP Morgan AI Research\\
London, UK\\
\texttt{benjamin.x.evans@jpmorgan.com} \\
\And
Leo Ardon \\
JP Morgan AI Research\\
London, UK\\
\And
Sumitra Ganesh \\
JP Morgan AI Research\\
New York, USA\\
}

\usepackage{amsmath}
\usepackage{amsfonts}

\DeclareMathOperator{\DKL}{\text{D}_\text{KL}}

\usepackage[capitalise]{cleveref}
\usepackage[font=small]{caption}
\usepackage[font=small]{subcaption}
\usepackage{booktabs}
\usepackage{paralist}

\begin{document}

\maketitle

\begin{abstract}
Modelling bounded rational decision-making through information constrained processing provides a principled approach for representing departures from rationality within a reinforcement learning framework, while still treating decision-making as an optimization process. However, existing approaches are generally based on Entropy, Kullback-Leibler divergence, or Mutual Information. In this work, we highlight issues with these approaches when dealing with ordinal action spaces. Specifically, entropy assumes uniform prior beliefs, missing the impact of a priori biases on decision-makings. KL-Divergence addresses this, however, has no notion of "nearness" of actions, and additionally, has several well known potentially undesirable properties such as the lack of symmetry, and furthermore, requires the distributions to have the same support (e.g. positive probability for all actions). Mutual information is often difficult to estimate. Here, we propose an alternative approach for modeling bounded rational RL agents utilising Wasserstein distances. This approach overcomes the aforementioned issues. Crucially, this approach accounts for the nearness of ordinal actions, modeling "stickiness" in agent decisions and unlikeliness of rapidly switching to far away actions, while also supporting low probability actions, zero-support prior distributions, and is simple to calculate directly.
\let\thefootnote\relax\footnotetext{\textbf{Extended Abstract}: Accepted at RLDM 2025, Dublin, Ireland.}
\end{abstract}


\begin{table}[!htb]
\centering
\begin{tabular}{@{}ccccc@{}}
\toprule
                        & \multicolumn{3}{c}{Metric}                     \\ \midrule
                        & \textbf{Entropy} & \textbf{KL*} & \textbf{W} \\ 
\textbf{Uniform Prior}  & \includegraphics[width=2cm]{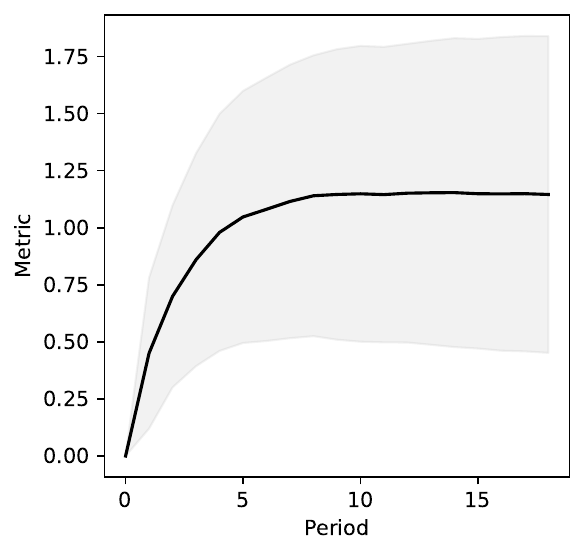} & \includegraphics[width=2cm]{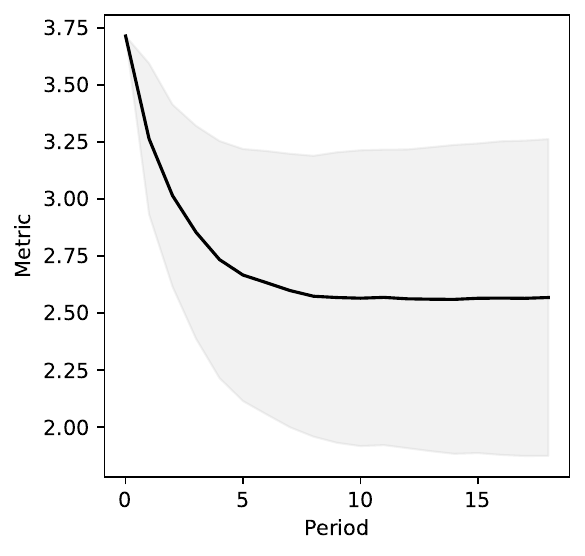} & \includegraphics[width=2cm]{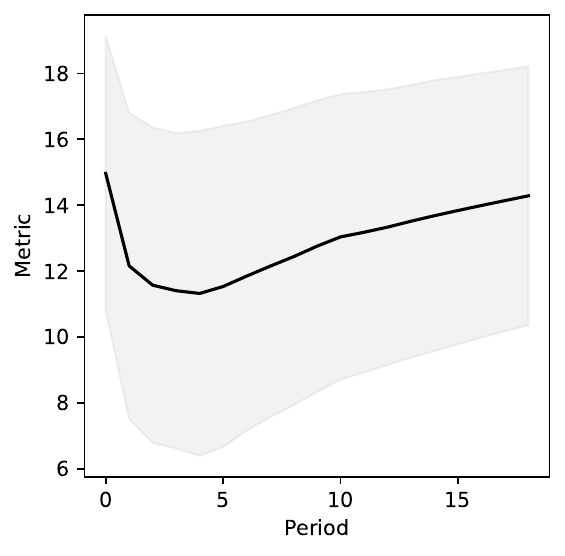} \\ 
\textbf{Previous Prior} & \includegraphics[width=2cm]{plots/entropy.pdf} & \includegraphics[width=2cm]{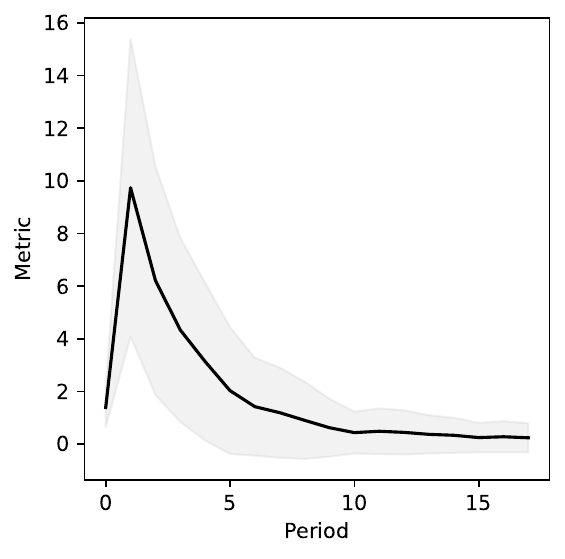} & \includegraphics[width=2cm]{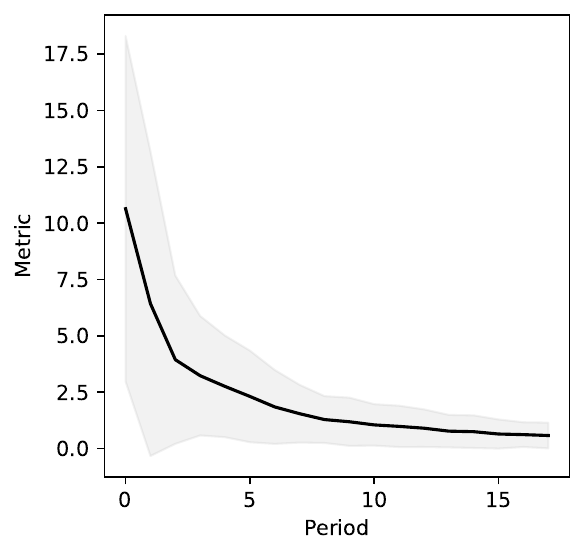} \\ 
\textbf{Optimal Prior}  & \includegraphics[width=2cm]{plots/entropy.pdf} & \includegraphics[width=2cm]{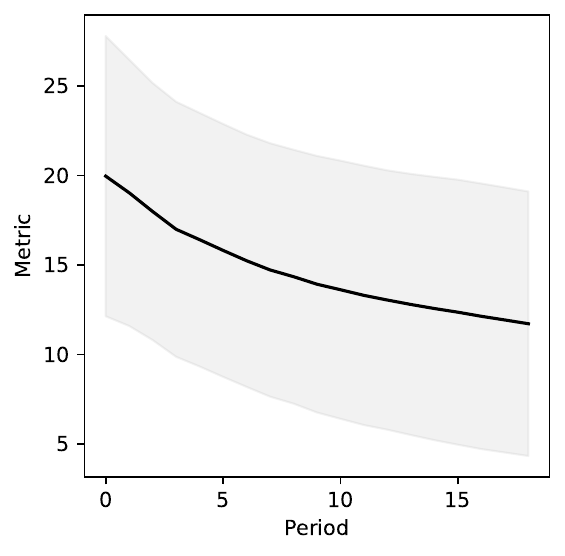} & \includegraphics[width=2cm]{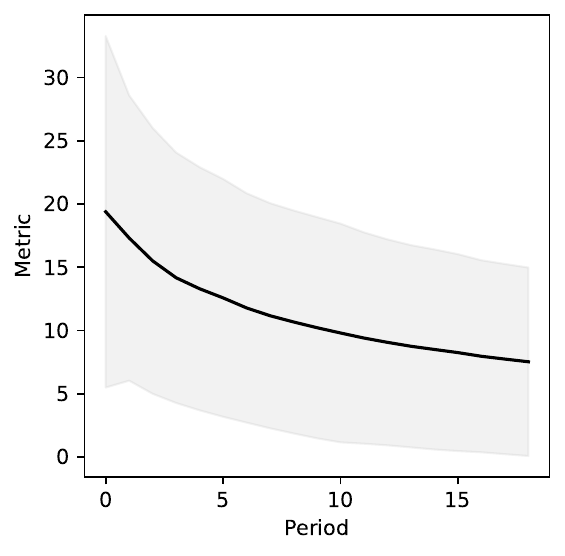} \\ 
\bottomrule
\end{tabular}
\caption{Metric across different prior beliefs. Note that KL is generally infinite for the previous and optimal priors, so we use a modified version KL*, which assigns a low probability to all zero probability actions to keep the metric finite.}\label{tblMetrics}

\end{table}

\startmain 

\section{Introduction}
Reinforcement Learning (RL) algorithms have achieved notable success in approximating optimal decision-making in complex sequential environments. However, when modeling human-like decision-making to simulate real-world behaviors (e.g., in traffic, markets), most prevailing methods assume perfectly rational agents. This assumption can be overly restrictive, failing to capture critical dynamics inherent in real-world systems \cite{evansAAMAS}. To address departures from strict rationality, various approaches have been proposed to model bounded rationality in RL, based on utility maximization under information-processing constraints. While promising, existing methods face limitations such as neglecting prior beliefs, imposing rigid forms of priors, ignoring action geometries, and computational challenges in their estimation. To overcome these issues, we propose a novel framework incorporating a Wasserstein distance-based constraint. In this extended abstract, we outline the concept and present motivating examples behind this idea.

\section{Background and Related Work}

Behavioural economics has developed more realistic models of decision-making than the traditional \textit{homo economicus} perfectly rational agent. Instead, these models operate under bounded rationality. While there are many different perspectives on bounded rationality \cite{kahneman2013perspective,gigerenzer2020bounded}, and the causes, here we focus on one particular representation that abstracts away specific causes, simply representing bounded rationality as decision-making under processing constraints \cite{ortega2013thermodynamics,evans2021maximum}.

Quantifying information processing costs in a generalised manner is desirable, as this enables compatibility with existing optimisation algorithms. This treatment abstracts the underlying causes of such constraints, allowing a focus on learning behaviour without necessitating an in-depth understanding of the specific psychological factors at play. From an optimisation standpoint, this is advantageous, as the process remains independent of the particular details of how decisions are formulated \cite{sims2003implications}. For experimentalists, a general enough form still allows for encoding different behavioural biases.

\subsection{Representation}

The general RL formulation is as follows. A decision-maker (DM) seeks to maximise their discounted return based on per time step \textit{utility} $U$ by taking actions from their action space $a \in A$. Importantly, these DM may not act perfectly rationally, and instead may be \textit{satisficing}. The system is characterised by a state space $S$, and DM's possess a (potentially partial) observation of the current state $s$ and prior beliefs about their potential actions $q$ (a probability distribution over the action space). The behaviour of the DM is governed by their policy $\pi$, which is a mapping from states to a distribution over actions. DM's act based on their policy $a \sim \pi$, receiving per timestep reward $U(a, s)$. Agents learn an (approximately) optimal policy $\pi_i^*$ that maximizes their expected lifetime return:
\begin{equation}\label{eqOriginal}
\pi_i^*(a|s_i) = \max_{\pi_i}\mathbb{E}_{\pi_i}\left[\sum_{t=0}^\infty\gamma^t U(a_t|s_{i,t}) \right]
\end{equation}

However, to model departures from perfect utility maximization, an alternative approach applies some form of information processing constraint to this maximization process, modelling limitations in reasoning capacities:
\begin{equation}\label{eqConstrained}
\begin{aligned}
\max_{\pi}\mathbb{E}_{\pi_i}\left[\sum_{t=0}^\infty\gamma^t U(a_t | s_{i,t}) \right] && 
\text{subject to} \quad & I(\pi_i, s_{i,t}, q_i) < \Bar{I}
\end{aligned}
\end{equation}
where agents maximise $U$ while adhering to a constraint $\Bar{I}$ on their processing costs $I$. Using a Lagrange multiplier, \cref{eqConstrained} can be reformulated as the maximization of a modified reward:
\begin{equation}\label{eqRewardMod}
    \pi_i^\lambda(a|s_i) = \max_{\pi_i}\mathbb{E}_{\pi_i}\left[\sum_{t=0}^\infty\gamma^t \left( U(a_t | s_{i,t}) - \lambda I(\pi_i, s_{i,t}, q_i) \right)\right]
\end{equation}

which importantly permits the same general representation as \cref{eqOriginal}, with a regularised utility function to model various departures from rationality based on the function $I$ (discussed in the following section) and prior beliefs $q$. These prior beliefs $q$ (also called "magnets" \cite{sokota2023} or "anchors" \cite{jacob2022modeling}) may change throughout training and inference (e.g. with updated information) and can take many forms, for example, demonstrating bias towards specific actions, encoding heuristics, averaging over past decisions, or preferring historically well-performing actions, allowing an additional form of bounded rationality when $I$ accounts for $q$. This constrained representation is beneficial, as it enables the utilization of any existing RL algorithm with minimal modifications to the loss function or optimization process \cite{evansAAMAS}.

\subsection{Existing information costs}\label{secLimitations} 

\begin{figure}[h!]
    \centering
    \begin{minipage}{0.45\textwidth}
        \centering
        \centering
     \begin{subfigure}[b]{0.3\textwidth}
         \centering
         \includegraphics[width=\textwidth]{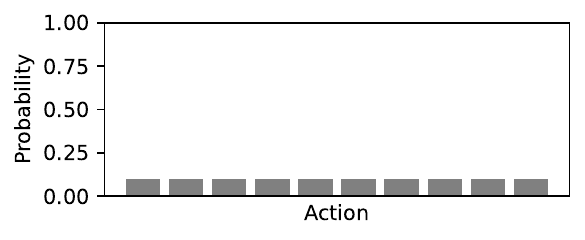}
         \caption{Low Entropy}
     \end{subfigure}
     \begin{subfigure}[b]{0.3\textwidth}
         \centering
         \includegraphics[width=\textwidth]{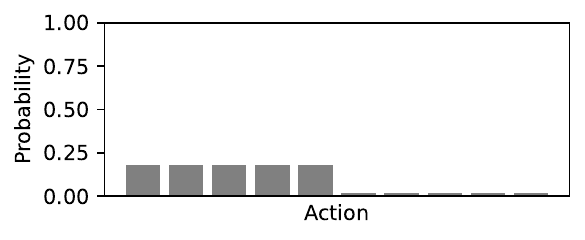}
         \caption{Mid Entropy}
     \end{subfigure}
     \begin{subfigure}[b]{0.3\textwidth}
         \centering
         \includegraphics[width=\textwidth]{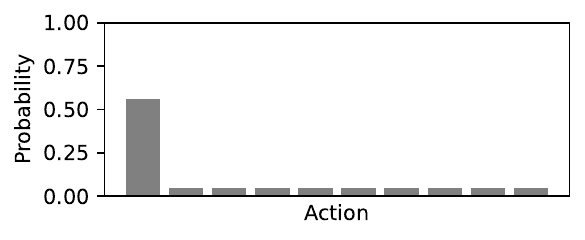}
         \caption{High Entropy}
     \end{subfigure}
        \caption{Entropy. Examples of various levels of entropy for different policies (independent of any prior beliefs)}
        \label{figEntropy}
    \end{minipage}
    \hfill
    \begin{minipage}{0.45\textwidth}
\begin{subfigure}[b]{0.3\textwidth}
         \centering
         \includegraphics[width=\textwidth]{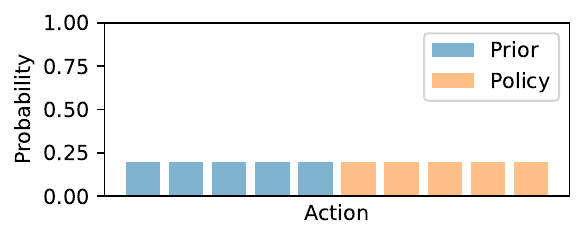}
         \caption{Example 1}
     \end{subfigure}
     \begin{subfigure}[b]{0.3\textwidth}
         \centering
         \includegraphics[width=\textwidth]{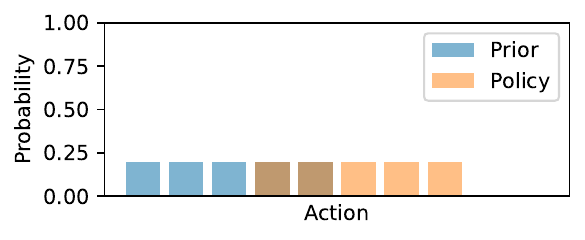}
         \caption{Example 2}
     \end{subfigure}
     \begin{subfigure}[b]{0.3\textwidth}
         \centering
         \includegraphics[width=\textwidth]{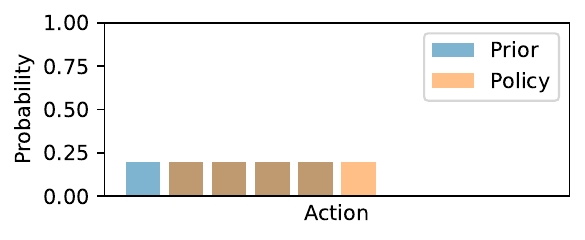}
         \caption{Example 3}
     \end{subfigure}
        \caption{KL examples with infinite divergence. Despite some of these policies seemingly being "closer" to the prior, all three have KL $=\infty$.} 
        \label{figKL}
    \end{minipage}
\end{figure}

By modifying $I$, we can model various forms of processing costs and capture different notions of bounded rationality. In this section, we examine common existing approaches, including entropy, KL-divergence, and Mutual Information.

\paragraph{Entropy}

An entropy constraint is one prominent approach for relaxing the strict, perfectly rational assumption, restricting deviations from uniform behaviour. For example, this is done in Quantal Response Equilibrium, which allows deviations from optimal responses and permits erroneous play. This constraint can be represented based on an information processing cost: $I_\text{Entropy} =  H = - \sum_{a \in A} \pi(a | s) \log \pi(a | s)$
which has multiple applications in RL \cite{sokota2023}.

However, much research has shown the usefulness of incorporating arbitrary prior beliefs $q$ (not just uniform) for better capturing realistic decision-making \cite{evans2021maximum}, motivating extensions that measure the divergence from an arbitrary prior distribution based on the Kullback-Leibler (KL) divergence $\DKL$ \cite{ortega2013thermodynamics}.

\paragraph{KL Divergence}

To better model human-like decisions and account for the impact of prior beliefs $q$ on decision-making, \cite{evansAAMAS} proposes a KL-based approach using the following information processing costs:
\begin{equation}\label{eqKL}
    I_{\DKL} = \DKL(\pi \parallel q) = \sum_{a \in A} \pi(a|s) \log \frac{\pi(a|s)}{q(a|s)}
\end{equation}
to constrain $\pi_i$ from diverging too far from agents' prior beliefs $q$ at each state, limiting their strategic abilities. When prior beliefs are uniform $q(a|s) = c$, \cref{eqKL} is equivalent to enforcing an entropy constraint (up to some constant), as 

\begin{equation}
    \sum_{a \in A} \pi(a|s) \log \frac{\pi(a|s)}{q(a|s)} 
     = \sum_{a \in A} \pi(a|s) \log \pi(a|s) -  \sum_{a \in A} \pi(a|s) \log c \\
      = -H + C \\
\end{equation}

However, in general cases, KL quantifies the divergence from \textit{arbitrary} prior beliefs, encoding different behavioral biases.

\paragraph{Mutual Information}

Finally, \cite{sims2003implications} proposes rational inattention (RI), which is based on Mutual Information, and this has been incorporated into RL in \cite{mu2022modeling}. MI is defined over the joint probabilities as:
$
   I_\text{MI} = - \sum_{a \in A} p(a,s_i) \log \frac{p(a,s_i)}{p(s_i)p(a)}
$
which has a dependence on the unconditional action probability $p(a)$ which generally must be solved with approximation techniques \cite{evans2021maximum}. For this reason, we focus our attention primarily on the alternative two approaches above due to their ability to be computed directly, as they only depend on the conditional probabilities directly given by the policies.


\textit{Limitations} Each of the above measures sufferers from their own limitations, including entropy not accounting for priors, KL going to infinity under many different configurations of priors, mutual information being challenging to compute, and none of the metrics accounting for the geometry or nearness of ordinal actions.


\section{Proposed Approach}

In order to overcome the aforementioned limitations in modeling bounded rationality in RL, in this section, we propose a novel RL approach based on Wasserstein distances.

\paragraph{Wasserstein metric}

We now define the Wasserstein distance $W$ (also known as the Kantorovich–Rubinstein metric or Earth Mover’s Distance) between two discrete probability distributions, the policy $p$ and prior beliefs $q$. While $W$ can also be defined on continuous distributions, we focus on the discrete case in this work. First, we quantify a distance \( d(a_i, a_j) \) between two actions \( a_i \) and \( a_j \) in the action space \( A \). We use the absolute distance \( d(a_i, a_j) = |i - j| \) for simplicity. For instance, the distance between actions 4 and 5 is \( d(4, 5) = 1 \). If there is no natural notion of distance between actions (e.g., for non-ordinal actions), we could instead use a fixed distance \( d(a_i, a_j) = D, \forall a_i, a_j \in A \), but generally, Wasserstein distances make the most sense under ordinal actions. Likewise, if there is a larger shift in agent perception required when moving between actions, for example, moving past some decision boundary, e.g. going from a positive to a negative action, larger distances could be assigned when crossing this boundary to represent the cognitive shift. 

We then construct a cost matrix \( C \), where \( C_{i,j} = d(a_i, a_j)^n \), for a chosen order \( n \) (e.g., \( n = 1 \) or \( n = 2 \)). The transport plan matrix \( T \) measures the cost of moving between a prior belief and a policy, and must satisfy the following constraints:
\begin{inparaenum}
    \item \( T_{i,j} \geq 0 \) (non-negativity),
    \item \( \sum_{a_i \in A} T_{i,j} = q(a_j) \) (supply constraint), and
    \item \( \sum_{a_j \in A} T_{i,j} =p(a_i|s) \) (demand constraint).
\end{inparaenum}

The Wasserstein distance is then defined as the optimal transport plan for this move from the prior to the policy:
\[
I_W = \min_T \sum_{i \in A} \sum_{j \in A} C_{i,j} T_{i,j},
\]
subject to the identified constraints. $I_W$ has several desirable properties, which we highlight in the experiments section, including the incorporation of prior beliefs, defined even on varying support, efficient to compute, and incorporating the geometry of the action space allowing for quantifying distances among actions, something not considered in KL-divergence or existing approaches. Additionally, $I_W$ is symmetric and satisfies the triangle inequality.

As discussed above, constrained decision-makers seek to maximise \cref{eqRewardMod}, i.e.: $$
    \pi_i^\lambda(a|s_i) = \max_{\pi_i}\mathbb{E}_{\pi_i}\left[\sum_{t=0}^\infty\gamma^t \left( U(a_t | s_{i,t}) - \lambda I_W(\pi_i, s_{i,t}, q_i) \right)\right]
$$

\section{Motivating Experiments}

To better motivate the chosen representation, we analyze a behavioural economics environment involving actual human participants and how their (inferred) polices evolve over time. We then use the discussed measures to quantify divergences from prior beliefs, showing where the proposed approach may be helpful.

\paragraph{Repeated public goods game}

We focus on a repeated public goods game (PGG) with experimental data from \cite{burton2015payoff}. In the PGG, DM's are given $40$ tokens and must decide how much of their tokens to contribute to a pool of public resources. Contributions to the public resource are multiplied by $1.6$ and dispersed equally to the players at the end of the round. In the experiments of \cite{burton2015payoff}, games are repeated for 20 rounds in groups of size 4.  The marginal-per-capita return for each unit contributed is 0.4; as this is $<1$, the strictly dominant rational strategy is for all players to thus contribute $0$. 

\begin{figure}[!htb]
     \centering
    \begin{subfigure}[b]{.25\textwidth}
         \centering
         \includegraphics[width=.95\textwidth]{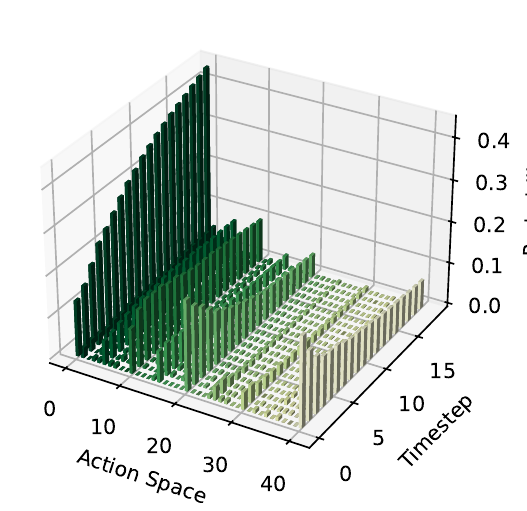}
         \caption{Average policy evolution}
     \end{subfigure}
     \begin{subfigure}[b]{.65\textwidth}
         \centering
         \includegraphics[width=.95\textwidth]{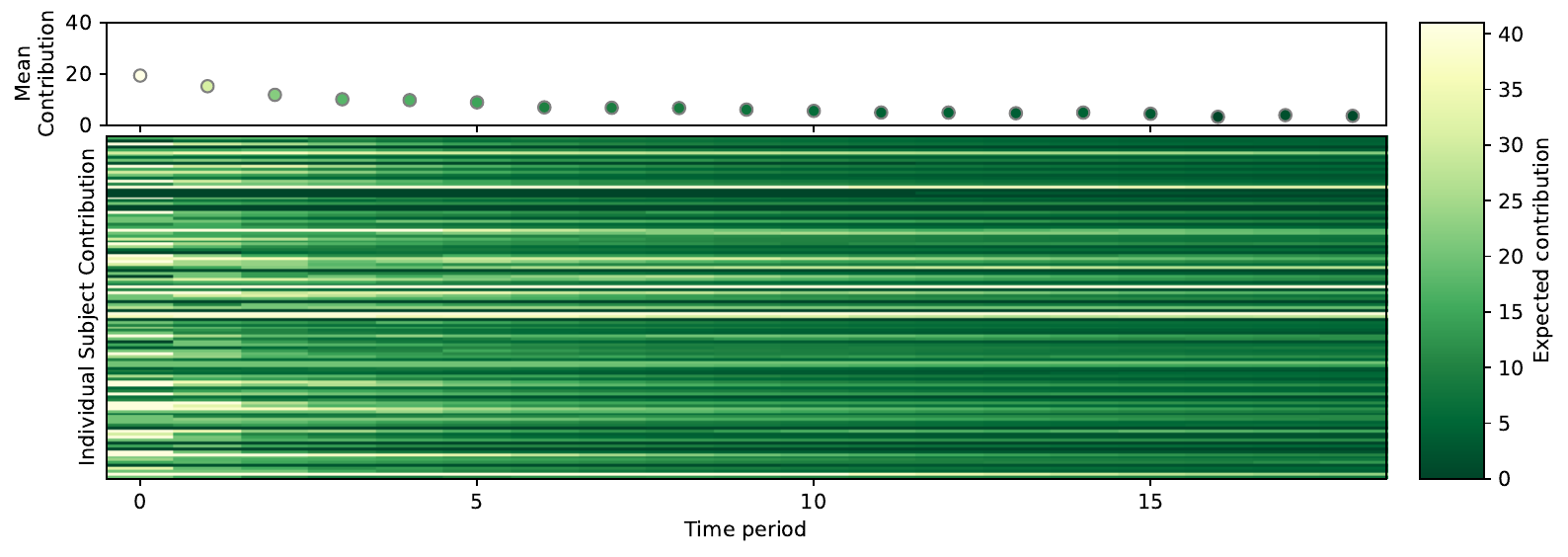}
         \caption{Overall contributions}
     \end{subfigure}
     \hfill
     \begin{subfigure}[b]{0.3\textwidth}
         \centering
    \includegraphics[width=\textwidth]{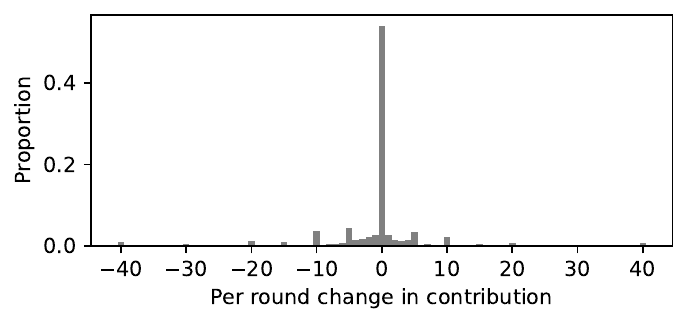}
         \caption{Change in contributions}
          \label{figPggChange}
     \end{subfigure}
     \begin{subfigure}[b]{0.3\textwidth}
         \centering
         \includegraphics[width=\textwidth]{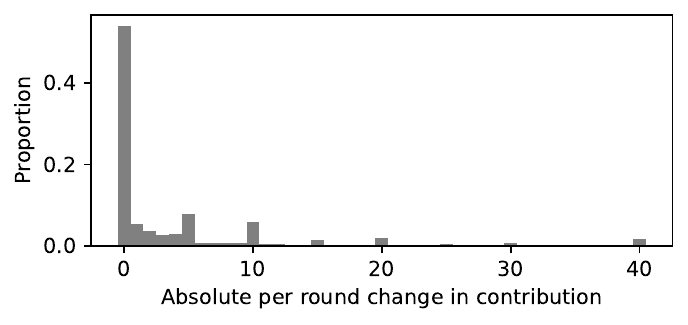}
         \caption{Absolute change in contributions}
         \label{figPggAbs}
     \end{subfigure}
     \begin{subfigure}[b]{.17\textwidth}
         \centering
         \includegraphics[width=.8\textwidth]{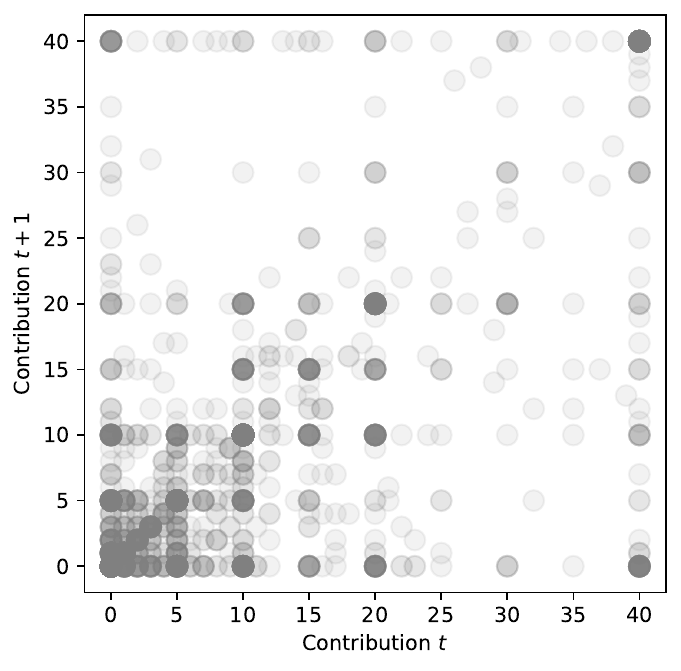}
    \caption{Pairwise changes}\label{figPairwise}
     \end{subfigure}
     \begin{subfigure}[b]{.17\textwidth}
         \centering
         \includegraphics[width=.8\textwidth]{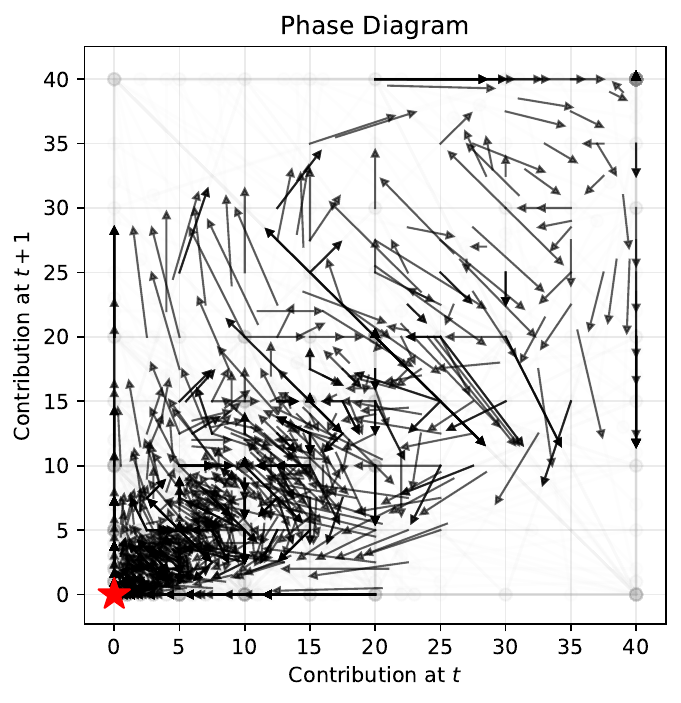}
         \caption{Phase Diagram}\label{figPhase}
     \end{subfigure}
     
        \caption{Public Goods Game, with real experimental data from \cite{burton2015payoff}}
        \label{figPgg}
\end{figure}

\subsection{Decisions}

To understand the rationality of decision-makers and also their willingness to make large changes in their decisions, we visualize the aggregated views of their contributions in \cref{figPgg}. While over time, there is a trend towards the rational choice (contributing $0$), we can see an apparent stickiness in their decisions, with the vast majority of decisions only changing their expected contribution by less than 5 each timestep, as well as peristing sub-rational choices.  

These changes are not only explained by the convergence towards the rational choice, as this is relatively symmetric for both decreases (e.g. approaching rationality) \textbf{and} increases in contributions (furthering from rationality), see \cref{figPggChange}. There are various proposed explanations for this, but the key is that these departures follow a clear bias towards previous decisions, as demonstrated in both \cref{figPggChange,figPggAbs}. \footnote{Additionally, we can see that there are peaks at prominent numbers (e.g. 5, 10, etc.), indicating the well-known prominent number bias, which could also be encoded in prior beliefs here or in the distance function from/to these prominent numbers.} We further confirm this by analyzing the per timestep change in contributions in \cref{figPairwise}, and the phase diagram of these changes in \cref{figPhase}.


\subsubsection{Prior beliefs and inferred Policies}

While we do not know the exact mental policies decision-makers were using or the prior beliefs of players in this game, only their sequentially revealed decisions, in \cref{tblMetrics}, we explore various priors and assume DM's policies are just the historical averages of the contributions they have played. We consider three different priors: uniform priors, previous timesteps policy (historical average, as above), and optimal priors. Uniform priors assign equal probability $\frac{1}{41}$ to each action $0\dots40$, previous timesteps policy is just the historical policy at $t-1$, and optimal prior is the Dirac delta function with all probability mass situated at the rational choice of 0, i.e., $p(0|\dots) =1$.

When using the different distance measures, \cref{tblMetrics} reveals some of the limitations discussed in \cref{secLimitations}, showing the benefits of the proposed approach. Entropy does not change under varying prior beliefs, meaning we can not model the influence of priors on resulting decisions. KL is infinite for previous and optimal priors, necessitating a modification assigning a low probability to all events. However, even with this modification, KL ends up exploding quite rapidly in the early periods due to the instabilities of logs of small numbers, making optimization difficult and potentially misleading. In contrast, the proposed Wasserstein-based approach is well-behaved under the three different circumstances, demonstrating that this provides a suitable alternative for modelling realistic human decision-making with RL.

\section{Conclusion}
In this work, we present an approach for modeling realistic decision-making within a RL framework, considering the geometry of action spaces. This approach leverages the Wasserstein distance between a DM's policy and their prior beliefs. We motivate its use by analyzing actual experiments with human participants, demonstrating that Wasserstein distance serves as a natural constraint for bounded rational decision-making. This extended abstract lays the groundwork for future exploration more complex RL environments, as well as ideas for improved efficiency of calculating the transport matrix, while highlighting the suitability and effectiveness of the proposed idea based on empirical economic studies.

\paragraph{Disclaimer}
{\footnotesize
This paper was prepared for informational purposes by the Artificial Intelligence Research group of JPMorgan Chase \& Co. and its affiliates ("JP Morgan'') and is not a product of the Research Department of JP Morgan. JP Morgan makes no representation and warranty whatsoever and disclaims all liability, for the completeness, accuracy or reliability of the information contained herein. This document is not intended as investment research or investment advice, or a recommendation, offer or solicitation for the purchase or sale of any security, financial instrument, financial product or service, or to be used in any way for evaluating the merits of participating in any transaction, and shall not constitute a solicitation under any jurisdiction or to any person, if such solicitation under such jurisdiction or to such person would be unlawful. © 2025 JPMorgan Chase \& Co. All rights reserved.
}

\vspace{-3mm}
\footnotesize
\bibliographystyle{ieeetr}
\bibliography{bib}

\end{document}